\documentclass[10pt,twocolumn]{article}

\usepackage[letterpaper,top=0.66in,bottom=0.72in,left=0.66in,right=0.66in,
            columnsep=0.25in,headsep=0.18in]{geometry}
\usepackage{newtxtext,newtxmath}

\usepackage{amsmath,amssymb}
\usepackage{graphicx}
\usepackage{booktabs,tabularx}
\usepackage{microtype}
\usepackage{authblk}
\usepackage{caption}
\usepackage{natbib}
\usepackage[hidelinks]{hyperref}

\hypersetup{
  pdftitle={Unanimity Without Persuasion: A Single Round of Debate Erases the Disagreement That Verification Needs},
  pdfauthor={Yang Shu}
}
\title{Unanimity Without Persuasion: A Single Round of Debate Erases the Disagreement That Verification Needs}
\author[1]{Yang Shu\thanks{Corresponding author: \href{mailto:shuyang@zju.edu.cn}{shuyang@zju.edu.cn}.}}
\affil[1]{Zhejiang University, Hangzhou, China}
\date{\small Preprint --- August 2026}

\begin{document}

\maketitle

\begin{abstract}
A debate panel can become unanimous without becoming more correct. This is dangerous for downstream safeguards: a substituted verification ballot can change only narrow-margin votes, while richer arbiters lose disagreement as a natural targeting signal. We show that one debate round can erase that resource without requiring persuasion. Tracking a heterogeneous 7-judge panel through a blind round and three debate rounds on 600 code-correctness candidates, unanimity on a fixed cohort jumps from 39.5\% to 95.2\% in round 1 (93.1\% of the total collapse), while accuracy moves by less than one point and 96.3\% of verdict flips follow the displayed peer majority. An execution-based verification ballot corrects 8 of 2{,}037 pre-debate candidate--substitution instances but changes \emph{zero} in every later round; by round 3 every wrong decision is unanimous, erasing dissent that had flagged two-thirds of the panel's errors. Identical-cohort controls explain why: no-peer reconsideration reproduces 79.6\% of the collapse, real labels without reasoning reproduce 91.5\%, and random labels steer flips toward whatever they display; the full-debate condition adds 4.3 percentage points over labels only (clustered 95\% CI 0.7--8.1). The one-round collapse reproduces in two additional real runs and two fake-label seeds, remains under panel sizes 3--7, and appears in MATH-500. Parse failures concentrate on contested candidates ($p<0.001$), making attrition non-ignorable. The design implication is operational: verify before any second-pass evaluation or peer exposure, and never treat post-debate unanimity as independent evidence of reliability.
\end{abstract}

\section{Introduction}

Multi-agent debate is usually evaluated by its final accuracy. Yet downstream safeguards depend on another quantity: whether the panel preserves enough disagreement to expose and correct suspect decisions. In the standard recipe, several agents answer independently, see one another's answers and arguments, revise, and a majority vote decides \citep{du2024improving,liang2024encouraging,chan2024chateval}. The recipe inherits its appeal from the Condorcet intuition that pooling many better-than-chance voters yields reliability \citep{condorcet1785essai}, and its known weakness from the fact that LLM voters are far from independent: their errors are strongly correlated \citep{kohli2026nine,kim2025correlated}, so majorities can systematically crush correct minority opinions \citep{he2026minority}. A natural remedy is to intervene on the panel's decisions from outside the majority: \citet{he2026minority} study exactly when a minority opinion should overturn a debate panel's majority, and their results show how treacherous such interventions are -- their baseline arbiter, an LLM reading the full debate log, flips more decisions wrongly than correctly (flip precision 42.7\%, net $-1.37$pp), and only a deliberately conservative trained gate flips safely. Why is intervening on a post-debate panel so hard?

\begin{figure}[t]
\centering
\includegraphics[width=\columnwidth]{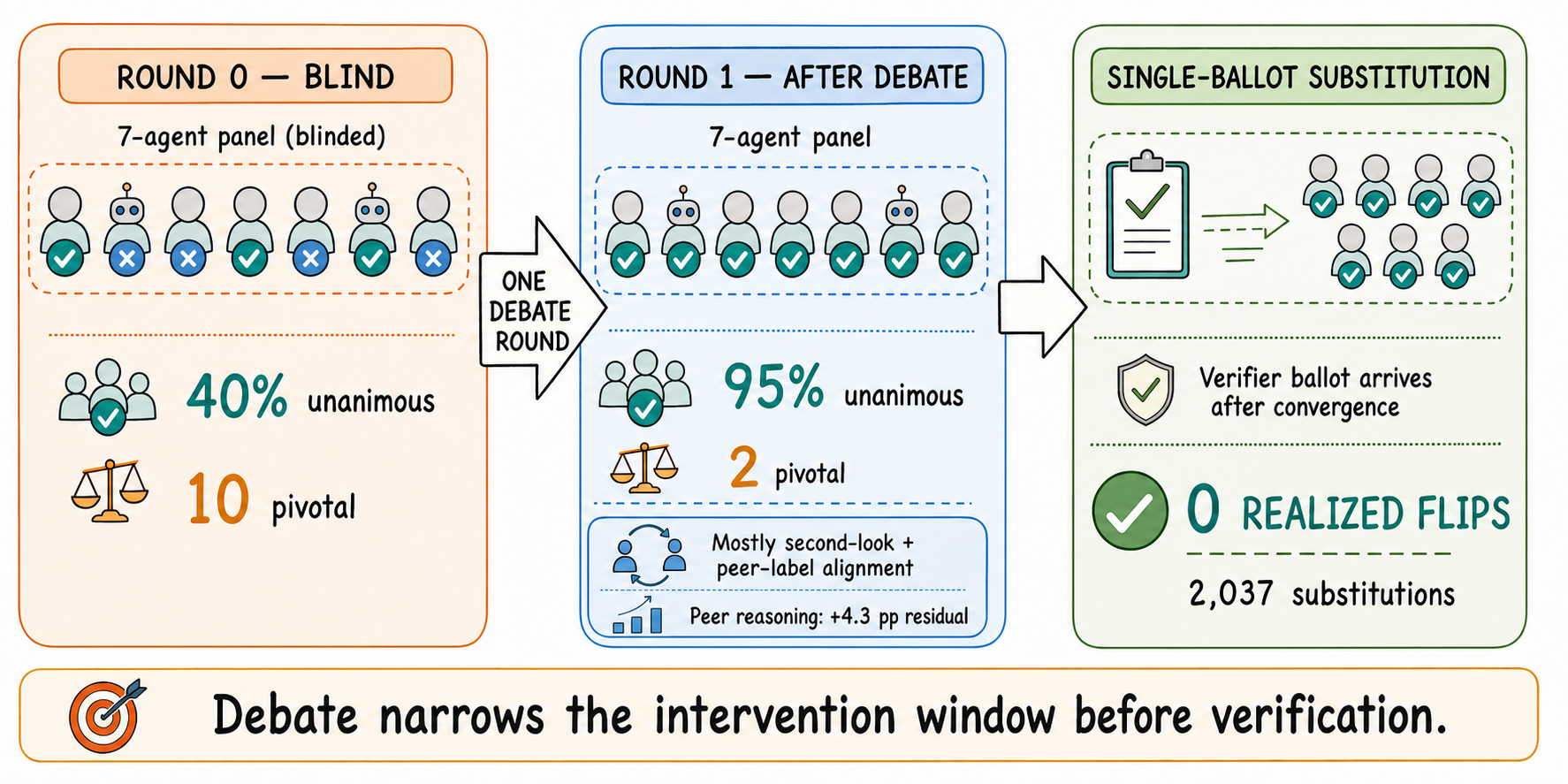}
\caption{Mechanism overview. Round metrics use the fixed $n{=}291$ cohort: one debate round takes unanimity from 39.5\% to 95.2\%, pivotal cases from 10 to 2, and execution-ballot substitution to 0/2{,}037 realized flips. The control callout uses the separate matched $n{=}280$ cohort (Section~5).}
\label{fig:teaser}
\end{figure}

Figure~\ref{fig:teaser} maps three claims: the paired margin trajectory, the separate matched-control decomposition, and the tested substitution outcome. Although two pivotal cases remain after round~1, replacing each judge in turn with the execution ballot changes no majority in 2{,}037 instances. This result is specific to that ballot and rule; arbitrary overrides remain possible.

Our answer starts from what substitution requires and richer overrides can exploit: \emph{disagreement}. The cleanest case is single-ballot substitution -- splicing one external vote, such as an execution-based verification signal, into the majority in place of one judge. An arithmetic fact (Fact~1) confines this intervention entirely to \emph{pivotal} queries, where the vote is within one ballot of flipping \citep{banzhaf1965weighted}. Stronger interventions, like \citeauthor{he2026minority}'s overriding arbiter, can change any decision, but must identify which decisions are likely wrong; panel disagreement is one natural targeting signal. The question therefore reduces to a measurable one: what does debate do to the panel's disagreement?

The answer, in our experiments, is that debate eliminates it -- fast. We track a panel of seven heterogeneous LLM judges (spanning seven providers) through a blind round and three debate rounds on 600 code-correctness candidates from HumanEval+ and MBPP+, where executable benchmark ground truth lets us score every panel decision deterministically. Before debate, the judges vote blindly and disagree often: only 25\% of candidates receive a unanimous verdict, and 16\% sit in the pivotal region. After a single round in which each judge sees its six peers' verdicts and up to 400 characters of reasoning per peer, disagreement is nearly gone. On the fixed cohort of candidates observable in every round ($n{=}291$; we return to why this conditioning is necessary below), unanimity jumps from 39.5\% to 95.2\% in round~1, then creeps to 98.6\% and 99.3\% in rounds~2--3: \emph{93.1\% of the total collapse happens in the first round} (problem-clustered bootstrap 95\% CI 88.9--96.8). Per-judge verdict flips fall from 11.8\% (round 0$\to$1) to 0.7\% and then 0.1\%; pivotal candidates drop from 10 to 2 to zero. The consequence for substitution is not an extrapolation -- we measure it. Substituting our execution-based verification signal for one judge changes the panel's round-0 decision in 8 of 2{,}037 candidate--substitution instances (two distinct pivotal candidates, every change a correction); applied to the round-1, round-2, or round-3 votes of the same candidates, the same substitution changes \emph{zero} instances per round. One round of debate reduces the verifier's realizable effect under substitution from small-but-real to exactly nothing.

Is this collapse at least \emph{earned} -- judges being persuaded by good arguments? A ladder of controls on an identical cohort says mostly no, and refines the mechanism. Candidate-independent random labels with template justifications reproduce 69\% of the unanimity jump (and 89.5\% of the flips they induce track the displayed random majority); peers' \emph{real} labels with all reasoning withheld reproduce 91.5\%; and -- the surprise -- a no-peer second-look prompt reproduces 79.6\%. Flip volume is nearly identical across all four conditions. The full-debate condition adds a small but detectable 4.3pp over labels only (clustered 95\% CI 0.7--8.1), while second-pass reconsideration and displayed labels explain most convergence. We connect the label-following component to classic conformity and anchoring \citep{asch1956studies,tversky1974judgment} and to LLM sycophancy \citep{sharma2024towards}.

A final finding began as due diligence and became evidence in its own right. Usable sample sizes shrink across rounds (600$\to$323$\to$300$\to$291) because some judge responses cannot be parsed and those candidates are dropped. We tested whether this attrition is random. It is not, at high significance: dropped candidates are pivotal at round~0 eight times as often as retained ones (31.1\% vs.\ 3.7\%, $\chi^2(3){=}151.9$, $p<0.001$), and the same pattern holds at every attrition point. Judges fail to even produce parseable output disproportionately on contested candidates. Methodologically, this means naive cross-sample comparisons (round-0 statistics on the full batch vs.\ round-1 statistics on survivors) overstate the collapse by about 13 points; every comparison in this paper is therefore made on fixed, paired samples. Because most parse failures trace to a single judge, we also verify that the entire collapse pattern survives removing that judge (a 6-judge panel with far less attrition, $n{=}298$: unanimity $55.4\%\to95.3\%\to98.7\%\to99.3\%$).

Our contributions:
(1)~a formal separation of intervention rules (substitution, addition, override), an exact decomposition of when \emph{substitution} can change decisions, and a direct round-by-round measurement of a real verification ballot's realizable effect -- corrective before debate, zero effect in every round after it (Table~\ref{tab:rounds});
(2)~a four-round ablation on a fixed cohort showing the collapse is front-loaded into round one (93\%, CI 89--97\%), flips are near-unanimously directed toward the displayed majority (96.3\%, CI 94--98\%), debate erases the panel's own error-flagging signal (accuracy barely moves while every wrong decision becomes unanimous), and the pattern survives excluding the most failure-prone judge;
(3)~a ladder of identical-cohort controls -- random labels, real labels only, and no-peer reconsideration -- decomposing the collapse into second-pass convergence (79.6\% of the jump), label alignment (to 91.5\%), and a small but detectable full-debate increment over labels only (4.3pp, clustered CI 0.7--8.1); and
(4)~evidence that response-parse failures concentrate on contested candidates, both a warning for debate experiment methodology (attrition is not ignorable) and an independent symptom of panel unreliability under disagreement.

\section{Related Work}

\paragraph{Multi-agent debate.} Debate among LLM agents has been proposed both as a capability amplifier \citep{du2024improving,liang2024encouraging} and as an evaluation device \citep{chan2024chateval,zheng2023judging}, with roots in the safety-via-debate program \citep{irving2018ai}. Empirical audits are increasingly skeptical: \citet{smit2024should} find debate protocols often fail to beat simpler ensembling; \citet{choi2025debate} disentangle debate from majority voting and find voting accounts for most of the gains; \citet{wu2025can}, in controlled logic puzzles, observe that majority pressure suppresses independent correction; \citet{zhu2026demystifying} attribute vanilla debate's underperformance to missing diversity and confidence calibration; while \citet{khan2024debating} show persuasion can help when debaters are strong. Our work is complementary to this line: rather than asking whether debate improves \emph{accuracy}, we quantify what it does to \emph{disagreement} -- the resource any external intervention on the panel needs -- and connect the two through a round-by-round measurement of a verification ballot's realizable effect. We are not aware of prior work reporting a judge panel's per-round vote-margin trajectory under debate, its control-based decomposition, or its consequence for single-ballot verification.

\paragraph{Intervening on debate outcomes.} \citet{he2026minority} is the direct motivation for this work. They study when a minority opinion should overturn a majority in multi-agent debate, training a conservative meta-classifier (flip precision 81.2\%) that decides when to override; their baseline -- an LLM arbiter reading the full debate log and deciding whether to overturn -- achieves higher recall but flips unsafely (flip precision 42.7\%) and yields a net $-1.37$pp effect. Their 22 meta-features include the vote margin (not in the top ten by importance), and no analysis of how debate itself reshapes the margin distribution is performed. Our work differs in both the intervention and the question. Their interventions are \emph{override} rules applied after debate; we distinguish the full spectrum (substitution, addition, override), directly measure a \emph{substitution}-type execution-based verification ballot round by round, and ask the upstream question their results presuppose: what does debate do to disagreement as an uncertainty and targeting signal? We therefore work in code correctness, where executable benchmark ground truth makes the relevant outcomes deterministically measurable. We offer a structural account of why the post-debate intervention problem is hard, not an audit of their system.

\paragraph{Correlated judges and where single ballots matter.} \citet{kohli2026nine} show LLM judge panels carry far fewer effective votes than nominal ones because errors are correlated, which motivates adding non-LLM evidence in the first place. The pivotal-voter notion we use to localize where such evidence can act is classical \citep{banzhaf1965weighted}; we claim no novelty for the arithmetic -- the empirical content of this paper is what debate \emph{does} to the margin distribution, which the arithmetic makes consequential.

\paragraph{Disagreement as an uncertainty signal.} A long line of work uses ensemble or sample disagreement as the confidence estimate driving selective prediction and abstention \citep{geifman2017selective,wang2023selfconsistency}; vote margins play the same role inside judge panels, and in our data disagreement is strongly informative at round 0 (panel accuracy 65.3\% on pivotal vs.\ 85.5\% elsewhere). Our results bear on this practice directly: the signal does not survive debate -- wrong decisions do not keep their dissent, they absorb it. We are not aware of prior measurements of the disagreement signal's fate \emph{under} debate.

\paragraph{Conformity, anchoring, sycophancy.} That individuals adopt majority positions under social exposure, even against the evidence of their senses, is among the oldest results in social psychology \citep{asch1956studies}, as is judgment anchoring on provided reference points \citep{tversky1974judgment} and group polarization toward consensus \citep{sunstein2002law}. LLMs exhibit a cognate failure mode as sycophancy toward stated user views \citep{sharma2024towards}, and a recent controlled study directly documents majority conformity in LLMs \citep{zhu2025conformity}. We do not claim to discover conformity. Our contribution is to account for its operational consequence inside a debate-based judge panel: matched controls attribute 79.6\% of the unanimity jump to a no-peer second look, take it to 91.5\% with bare real labels, and leave a small 4.3pp residual for the full-debate condition; we then connect that collapse to the verifier opportunity the panel loses.

\section{Setup: Panels, Margins, and Where a Ballot Can Matter}

\paragraph{Candidates and ground truth.} We use 600 (problem, solver) candidate solutions drawn from HumanEval+ and MBPP+ \citep{chen2021evaluating,austin2021program,liu2023your}, produced by 7 solver models of varied capability (Claude Opus, GPT-5, DeepSeek-V4-Pro, GLM-5, Kimi-K2.7-Code, Doubao-1.5-Lite, GPT-4.1-mini) so that correct and incorrect solutions occur naturally; the solver and judge rosters are disjoint, though some share providers. Ground truth is execution against each problem's full extended test suite in a sandboxed subprocess. An execution-based \emph{verification signal} -- fallible by construction -- is the pass/fail outcome on a fixed 20\% subset of test cases, chosen independently of model outputs; it misses bugs on untested inputs, a different failure mode from an LLM judge's susceptibility to plausible-looking code.

\paragraph{Why code correctness.} Every quantity our argument needs -- whether the majority is right, whether a unanimous verdict is wrong, whether a flip helped -- requires a deterministic reference, which extended execution supplies without human labels; the domain also cleanly separates failure modes (a judge is fooled by code that \emph{looks} right; a partial test suite misses bugs on inputs it does not run).

\paragraph{Judges and rounds.} Seven LLM judges from seven providers (GPT-4o-mini, Claude Haiku, DeepSeek-V4-Flash, Qwen3-max, GLM-5-Turbo, Kimi-K2.6, Doubao-Seed-1.6) score each candidate as correct or incorrect. \textbf{Round 0} is a blind, independent judgment: no judge sees any other's output, and none sees execution results. In \textbf{round $r\geq1$}, each judge is shown the six peers' round-$(r{-}1)$ verdicts and reasoning excerpts (up to 400 characters per peer) -- anonymized, in randomized order, so that no judge can weight peers by identity -- and asked to re-decide the same candidate. These case-specific excerpts distinguish full debate from the labels-only control, but do not isolate semantic persuasion from context length or presentation. All queries run at temperature 0 with the verdict required on the first line of the response; candidates for which any judge's response cannot be parsed into a verdict in some round are permanently dropped from that round onward (we analyze this attrition explicitly below).

\paragraph{Margins and the pivotal region.} For candidate $i$ let $s_i = \sum_j \mathbb{1}[\hat y_{ij} = 1] \in \{0,\dots,7\}$ be the number of votes \emph{labeling the candidate correct} (not the number of ground-truth-correct votes) and $m_i = |2s_i - 7|$ the \emph{margin}. $m_i{=}7$ is unanimity; $m_i{=}1$ (a 4--3 split) is \emph{pivotal}.

\paragraph{Three intervention rules, kept distinct.} An external signal $v_i \in \{0,1\}$ can enter the panel's decision in inequivalent ways, and their arithmetic differs: (i)~\emph{substitution} -- $v_i$ replaces one judge's ballot, keeping $k{=}7$; (ii)~\emph{addition} -- $v_i$ joins as an eighth ballot, which requires a tie-breaking convention for 4--4; (iii)~\emph{override} -- a policy may replace the panel's decision outright, as in \citeauthor{he2026minority}'s arbiter. This paper's measurements use substitution, the cleanest rule; we return to override qualitatively in the discussion. The structural fact for substitution is elementary:

\begin{quote}
\textbf{Fact 1.} \emph{Let a $k$-judge panel ($k$ odd) decide candidate $i$ by unweighted majority vote. Under substitution of exactly one ballot, the panel's decision can change if and only if $m_i = 1$.}
\end{quote}

\emph{Proof.} Substitution changes the tally $s_i$ by at most one. For odd $k$ the decision boundary lies strictly between $\lfloor k/2 \rfloor$ and $\lceil k/2 \rceil$, so a $\pm 1$ change crosses it only from $s_i \in \{\lfloor k/2\rfloor, \lceil k/2\rceil\}$, i.e.\ $m_i = 1$. $\blacksquare$

Fact 1 fixes the \emph{magnitude scaffolding} of substitution, not the sign. The expected accuracy change decomposes as
\begin{equation*}
\Delta A \;=\; P(m{=}1)\,\bigl[\,P(\text{fix} \mid m{=}1) - P(\text{break} \mid m{=}1)\,\bigr],
\end{equation*}
where \emph{fix} (\emph{break}) is the event that the substitution flips a wrong (right) majority, and expectations are taken jointly over candidates and the replaced judge (we evaluate all seven replacement positions and average). The pivotal share bounds how much effect is achievable; the signal's conditional quality on pivotal queries sets the sign. Outside $m{=}1$ the effect is identically zero -- neither help nor harm -- for any signal, oracle or garbage. We claim no novelty for the arithmetic (it is the classical pivotal-voter observation of \citealp{banzhaf1965weighted}, specialized to substitution); its role is to make the margin distribution the load-bearing quantity. What requires evidence is where panel errors live and what debate does to that distribution.

\paragraph{A motivating stratification.} On the 600 blind round-0 votes, panel errors concentrate near the boundary: majority-vote accuracy is 65.3\% on the 98 pivotal candidates against 85.5\% on the rest (82.2\% overall). Substituting the verification signal for one judge (averaged over all seven leave-one-out choices) improves pivotal-region accuracy by $+5.83$pp -- but the bootstrap 95\% CI, $[-1.1, +12.6]$, crosses zero at this sample size, and we report it as directional rather than established. The gain outside the pivotal region is $0.0000$, as Fact 1 requires. Nothing downstream depends on the significance of this round-0 gain: the paper's claims rest on the margin-distribution dynamics and the round-by-round substitution measurements of Section 4.5, to which we now turn.

\section{Debate Destroys Disagreement Within One Round}

\subsection{Attrition Is Not Random -- So All Comparisons Are Paired}

Usable candidates shrink across rounds: 600 at round 0, then 323, 300, 291 (and 416 in the fake-verdict condition of Section~5), because a candidate is dropped once any judge's response fails to parse. Before comparing rounds we ask whether the dropped candidates resemble the retained ones. They do not (Table~\ref{tab:attrition}). At every attrition point, dropped candidates are drastically more contested at round 0: at round 1, the dropped 277 candidates have a 31.1\% pivotal share and 9.0\% consensus share, against 3.7\% and 38.7\% for the 323 retained ($\chi^2(3){=}151.9$, $p<10^{-3}$); their round-0 panel accuracy is also far lower (70.8\% vs.\ 92.0\%). The pattern repeats at rounds 2, 3, and in the control condition ($\chi^2$ from 78.9 to 135.9, all $p<10^{-3}$).

\begin{table}[t]
\centering
\small
\begin{tabular}{@{}lrrrrr@{}}
\toprule
 & \multicolumn{2}{c}{$n$} & \multicolumn{2}{c}{Pivotal share (R0)} & \\
\cmidrule(lr){2-3}\cmidrule(lr){4-5}
Attrition point & ret. & drop. & ret. & drop. & $\chi^2(3)$ \\
\midrule
Round 1 & 323 & 277 & 3.7\% & 31.1\% & 151.9 \\
Round 2 & 300 & 300 & 3.7\% & 29.0\% & 135.9 \\
Round 3 & 291 & 309 & 3.4\% & 28.5\% & 134.7 \\
Round 1 (fake) & 416 & 184 & 9.6\% & 31.5\% & 78.9 \\
\bottomrule
\end{tabular}
\caption{Parse-failure attrition is strongly non-random: candidates dropped for unparseable judge output are far more often pivotal at round 0 than retained ones ($\chi^2$ over the four round-0 margin bins $\{1,3,5,7\}$, df$=3$; all $p<10^{-3}$). Retained/dropped round-0 panel accuracy differs correspondingly (e.g.\ 92.0\% vs.\ 70.8\% at round 1).}
\label{tab:attrition}
\end{table}

Two consequences. First, methodological: comparing full-batch round-0 statistics (25.0\% consensus) against survivor-only round-1 statistics (93.8\%) inflates the apparent collapse by roughly 13 points, because the survivors started out more agreeable. All cross-round comparisons below therefore use the \emph{fixed cohort} of 291 candidates usable in every round, and all cross-condition comparisons in Section~5 use identical-cohort intersections. Second, substantive: the panel's output discipline itself degrades on contested candidates -- an independent symptom that the contested region is where panel behavior is least trustworthy. The cost of the paired design is stated plainly: the fixed cohort over-represents easy candidates (39.5\% round-0 unanimity vs.\ 25.0\% in the full batch), so our collapse estimates describe the surviving subpopulation.

\paragraph{The collapse is not a single-model artifact.} Because most parse failures trace to one judge (Doubao-Seed-1.6; see Limitations for the full attribution), we recompute the entire trajectory on the 6-judge panel that excludes it. Attrition drops sharply (fixed cohort $n{=}298$ of 600), and the collapse is fully reproduced: unanimity $55.4\% \to 95.3\% \to 98.7\% \to 99.3\%$ across rounds, with per-judge flip rates $9.4\% \to 0.8\% \to 0.1\%$. Neither the front-loading nor the magnitude of the collapse depends on the panel's most failure-prone member.

\subsection{The Collapse Is Front-Loaded}

\begin{figure}[t]
\centering
\includegraphics[width=\columnwidth]{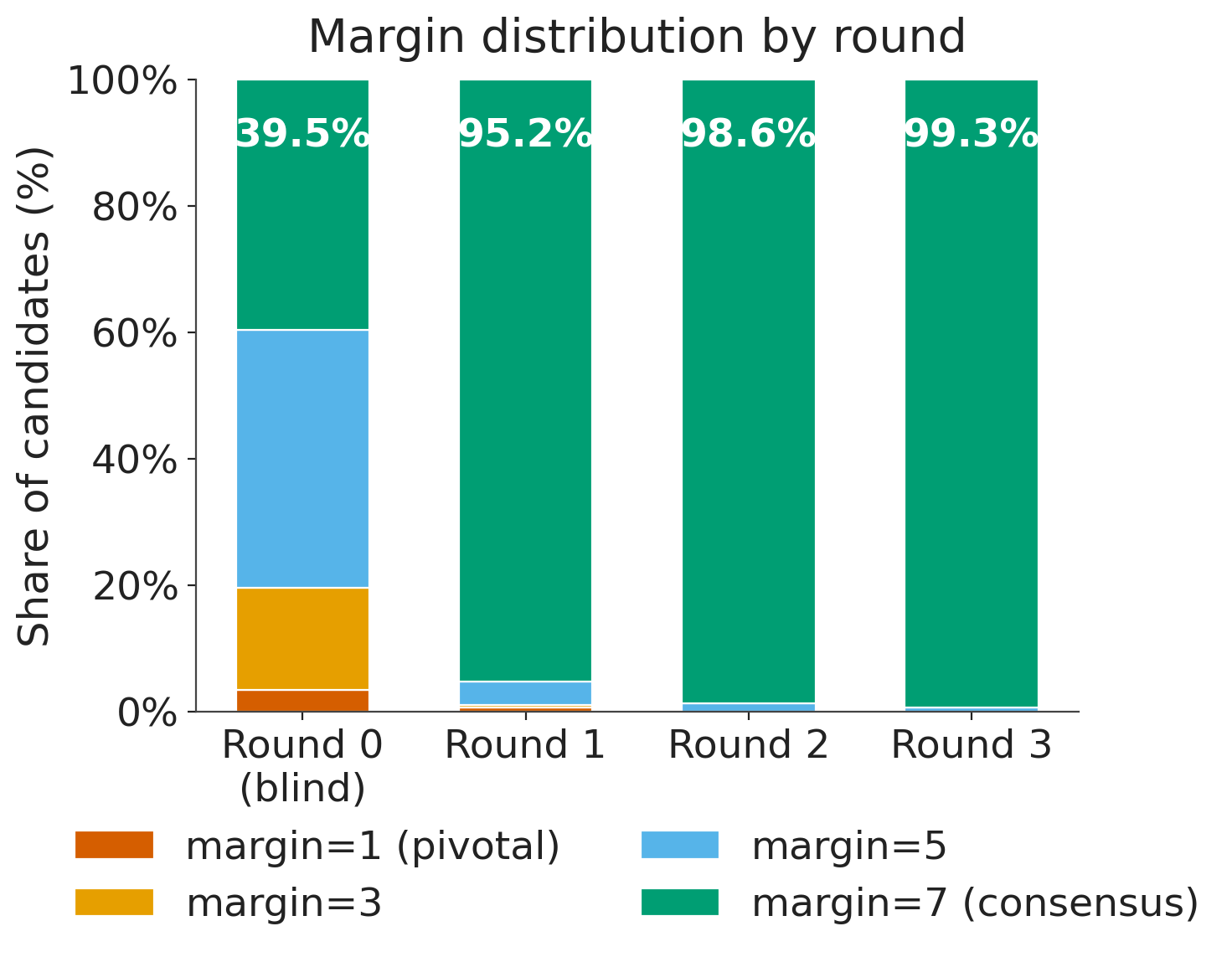}
\caption{Debate-induced consensus collapse on the fixed cohort ($n{=}291$ candidates usable in all rounds). The pivotal share (orange-red) falls from 3.4\% to 0.7\% after one round and to zero thereafter; unanimity (green) rises from 39.5\% to 95.2\% in round 1, then only to 98.6\% and 99.3\%. Essentially the entire effect of debate on disagreement is exhausted by the first exposure to peers' opinions (per-judge flip rates: Table~\ref{tab:rounds}).}
\label{fig:collapse}
\end{figure}

Table~\ref{tab:rounds} and Figure~\ref{fig:collapse} give the four-round trajectory on the fixed cohort. Unanimity rises $39.5\% \to 95.2\% \to 98.6\% \to 99.3\%$: of the total $+59.8$pp collapse over three debate rounds, $+55.7$pp -- 93.1\% (problem-clustered bootstrap 95\% CI $[88.9, 96.8]$) -- occurs in round 1. Pivotal candidates go $10 \to 2 \to 0 \to 0$. Per-judge flips (2{,}037 judge--candidate decisions per transition) drop by a factor of 16 between the first and second transitions and by another factor of 7 after that; by round 3, two flips remain out of 2{,}037. Debate does not gradually anneal the panel toward agreement; the first sight of peers' opinions does nearly all the work, and later rounds merely tidy a consensus that is already locked. (Interim nested samples collected during development showed the same qualitative pattern but are not independent replications; robustness rests instead on the clustered CIs, on the exclude-one-judge check of Section 4.1, and on the fact that the same front-loaded collapse appears in every 3-, 5-, and 6-judge sub-panel of the roster and at every verifier test-coverage level from 5\% to 50\%, Supplementary~D--E.)

\begin{table}[t]
\centering
\small
\setlength{\tabcolsep}{4pt}
\begin{tabular}{@{}lrrrrr@{}}
\toprule
 & & & & \multicolumn{2}{c}{Verifier subst.} \\
\cmidrule(lr){5-6}
Round & Pivotal & Unanimity & Flips & Changes & Net \\
\midrule
0 (blind) & 10 & 39.5\% & --- & 8 (8\checkmark) & $+0.39$pp \\
1 & 2 & 95.2\% & 11.78\% & 0 & $0.00$ \\
2 & 0 & 98.6\% & 0.74\% & 0 & $0.00$ \\
3 & 0 & 99.3\% & 0.10\% & 0 & $0.00$ \\
\bottomrule
\end{tabular}
\caption{Blind round plus three debate rounds on the fixed cohort ($n{=}291$). ``Flips'' is per-judge verdict changes vs.\ the previous round (of $291\times7{=}2{,}037$). ``Verifier subst.'': decision-changing candidate--substitution instances from splicing the execution ballot in place of each judge in turn (2{,}037 instances per round; the 8 round-0 changes are 2 unique candidates $\times$ 4 majority-side replacements, all corrections) and the net accuracy gain averaged over replacement positions.}
\label{tab:rounds}
\end{table}

\subsection{The Flips Are a One-Way Ratchet}

What kind of mind-changing produces this collapse? On the separate round-0/round-1 paired subset ($n{=}323$, larger than the four-round fixed cohort), we classify every one of the 276 individual verdict flips by its direction relative to what the flipping judge was actually shown: the majority verdict among its six peers' round-0 votes. Excluding 6 flips where the judge saw a 3--3 tie, \textbf{96.3\% of flips (260/270; clustered CI $[94.0, 98.4]$) moved toward the peer majority}. Debate-induced mind-changing is not bidirectional error correction in which arguments sometimes win against the crowd; it is almost pure absorption of dissenters into whatever majority happens to be on display. (Section 5.2 shows most of this flip \emph{volume} under the no-peer second-look prompt; what peers control is its direction.)

\begin{table}[t]
\centering
\small
\begin{tabular}{@{}lrr@{}}
\toprule
Judge & Real flips & Fake flips \\
\midrule
GLM-5-Turbo & 0.7\% & 1.7\% \\
Kimi-K2.6 & 2.3\% & 3.6\% \\
Claude Haiku & 7.3\% & 7.9\% \\
DeepSeek-V4-Flash & 9.2\% & 10.6\% \\
Qwen3-max & 13.9\% & 14.5\% \\
GPT-4o-mini & 24.1\% & 26.1\% \\
Doubao-Seed-1.6 & 26.1\% & 26.1\% \\
\bottomrule
\end{tabular}
\caption{Per-judge round-0$\to$1 flip rates under real and fabricated peer opinions, both computed on the identical $n{=}303$ candidates usable in both conditions. Rates span an order of magnitude (0.7\%--26.1\%) and the cross-condition ranking is almost perfectly preserved (Spearman $\rho = 0.99$, $p < 0.001$, bootstrap CI $[0.93, 1.0]$). Every judge flips at least as often in the fake condition, where the i.i.d.\ random labels confront it with an opposing majority more often -- so absolute rates are not comparable across conditions, only rankings.}
\label{tab:perjudge}
\end{table}

The propensity to be absorbed also varies enormously and consistently across models (Table~\ref{tab:perjudge}): on the identical candidates, per-judge flip rates span 0.7\% (GLM-5-Turbo) to 26.1\% (Doubao-Seed-1.6), and the ranking is almost perfectly preserved when the peers' opinions are replaced with random noise in Section 5's control (Spearman $\rho = 0.99$). With only seven models this is exploratory and may reflect baseline accuracy, calibration, or response style as well as conformity; it nevertheless motivates synthetic-peer probes as a descriptive screening tool.

\subsection{Confidence Explodes; Correctness Does Not}

If the flips were information exchange, accuracy should track confidence. It does not. On the fixed cohort, majority-vote accuracy moves from 91.8\% (round 0) to 92.4\% (rounds 1--3) -- two candidates corrected, none broken, a change indistinguishable from noise (exact McNemar $p{=}0.50$) -- while unanimity rises 60 points. The most consequential change is not in how often the panel is wrong but in \emph{how it is wrong}. At round 0 the panel's majority errs on 24 candidates, and 16 of those errors -- two-thirds -- occur on split votes: the disagreement itself flags them as unreliable. By round 3 the panel errs on 22 candidates, and \textbf{every single one is unanimous}. The count of unanimously wrong verdicts nearly triples (8 $\to$ 22), not because debate creates new errors, but because it converts errors that carried a visible dissent signal into errors carrying the panel's full confidence. Any downstream consumer that uses panel disagreement as an uncertainty estimate -- for routing, abstention, or triggering verification \citep{geifman2017selective,wang2023selfconsistency} -- loses its entire signal in one round.

On the same $n{=}323$ paired subset, ten of the twelve round-0 pivotal candidates reach margin $\geq 5$ within one round (majority outcome improves on two, degrades on none): debate is not harmful \emph{on} the contested region it can still see -- it eliminates that region as a distinguishable class.

\subsection{The Verifier's Realizable Effect, Round by Round}

Fact 1 plus a vanishing pivotal region predicts that a substituted verification ballot should lose all purchase after round 1. We test the prediction directly rather than leave it as an inference (Table~\ref{tab:rounds}, last columns): for each round's votes on the fixed cohort, we substitute the execution signal for each judge in turn ($291 \times 7 = 2{,}037$ candidate--substitution instances per round) and count decision changes against ground truth. At round 0 the substitution changes 8 instances -- two distinct pivotal candidates, each flipped by the four majority-side replacements, every change a correction -- for a net gain of $+0.39$pp averaged over replacement positions (clustered CI $[0.0, 1.0]$; concentrated in the pivotal region, where the within-region gain is the $+5.83$pp of Section 3). At rounds 1, 2, and 3, the same substitution changes \textbf{zero} instances per round. The verifier's realizable effect under substitution is not merely reduced by debate; it is annihilated, exactly as the margin distribution requires. We stress the rule-dependence of the sign: under substitution the post-debate effect is exactly zero, not negative. Negative net effects of the kind \citet{he2026minority} report arise under \emph{override}-type rules, which can corrupt any decision and therefore benefit from a reliable targeting signal; panel disagreement is one such signal, and its destruction is what we measure here.

\section{What Drives the Collapse? Label, Reasoning, and Second-Look Controls}

The collapse could still be benign: judges might be genuinely persuaded by peers' arguments, in which case convergence is information exchange working as intended. To separate protocol components -- case-specific peer-reasoning excerpts, bare verdict labels, and a no-peer second look -- we rerun round 1 on the same batch, from the same round-0 baseline, under three controlled conditions: (i)~\emph{fake peers}: verdict labels drawn independently of the candidate with content-free template justifications; (ii)~\emph{labels only}: the peers' \emph{real} round-0 verdict labels, with all reasoning withheld; and (iii)~\emph{no-peer reconsideration}: the judge is explicitly asked to take a second careful look, with no peer information. This last control measures repeated, reframed evaluation rather than an identical-prompt re-ask. Judges never see execution results in any condition.

\subsection{Random Labels Alone Reproduce Most of It}

\paragraph{Design.} In the fake-peers condition, each fake verdict is an independent draw reading \textsc{correct} with probability $0.82$ -- matching the marginal rate of \textsc{correct} verdicts in round 0, so the label distribution is not conspicuously abnormal -- drawn \emph{without any reference to the candidate}: the fake panel carries zero mutual information about the case. Each candidate gets one coherent fake panel (a fixed 7-slot verdict vector; every judge sees the other six slots, so any two judges see consistent opinions for the same fake peer). Justifications are fixed template sentences (``the solution appears to correctly implement the required logic\ldots'' / ``\ldots appears to have a logical error\ldots''). The draws are seeded and fully reproducible. If convergence requires persuasive, case-relevant argument, this condition should produce little collapse.

\begin{figure}[t]
\centering
\includegraphics[width=\columnwidth]{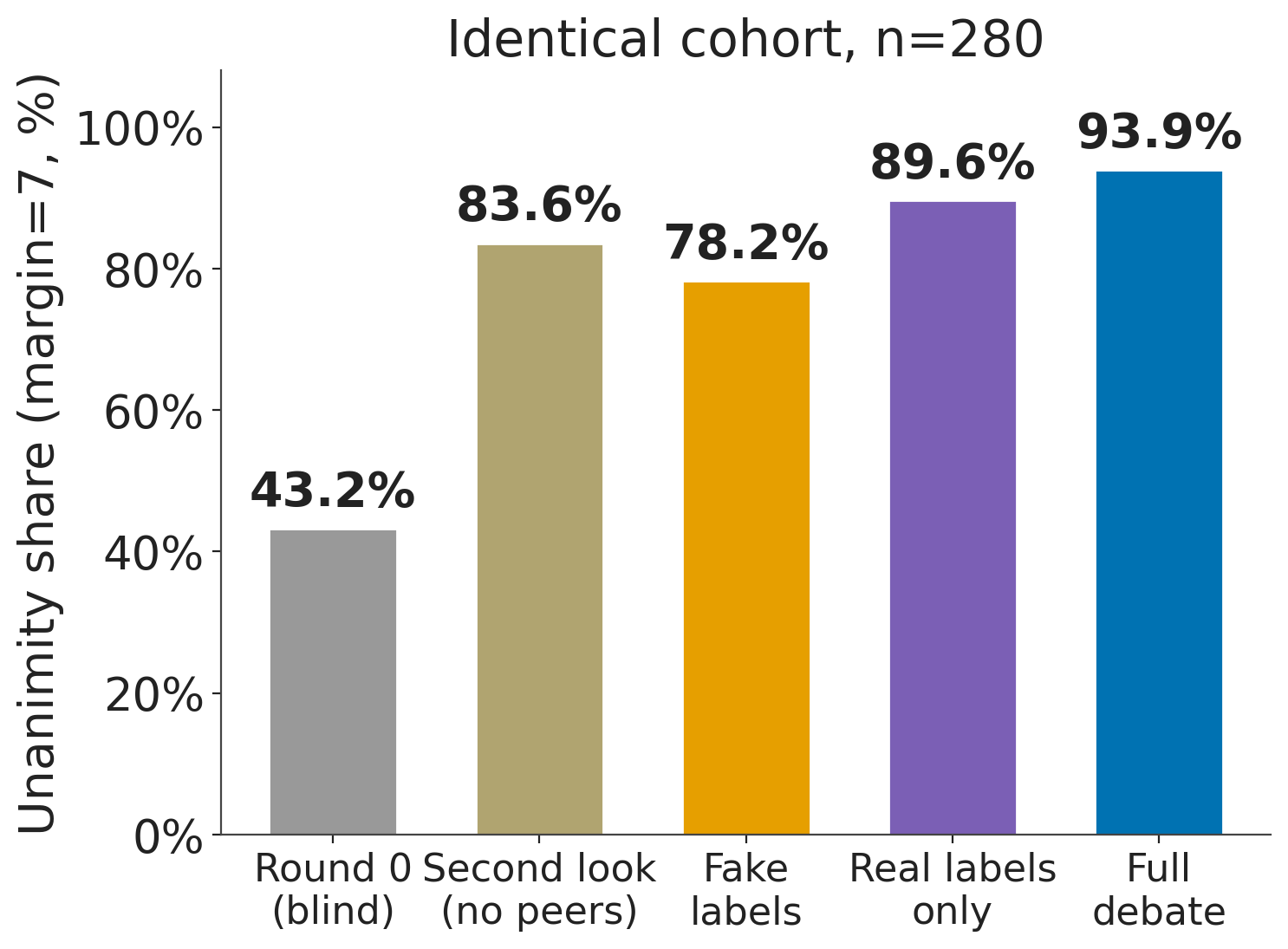}
\caption{Five conditions on the identical cohort of $n{=}280$ candidates. A no-peer second look reaches 79.6\% of full debate's unanimity jump; random labels steer flips but slow net convergence; real labels without reasoning reach 91.5\%; the full-debate condition adds 4.3pp over labels only. Flip volume is nearly identical in all four intervention conditions (10.4--11.8\%).}
\label{fig:control}
\end{figure}

\paragraph{Result.} On the 303 candidates usable in both conditions (Figure~\ref{fig:control}), round-0 unanimity is 40.3\%. Real peer opinions lift it to 93.7\% ($+53.5$pp). The fabricated ones lift it to 76.6\% ($+36.3$pp): \textbf{67.9\% of the real jump (problem-clustered bootstrap 95\% CI $[58.0, 76.6]$) is reproduced by verdict labels that carry no information about the case}. Because the fake draws are seeded, we can reconstruct exactly which random majority each judge saw and repeat the direction analysis of Section 4.3: \textbf{89.5\% of flips in the fake condition (410/458, excluding 14 shown ties) moved toward the displayed random majority} -- judges track whichever way the noise happens to point. (The i.i.d.\ labels lean \textsc{correct} and most candidates are in fact correct, so this herding is even mildly beneficial here -- a base-rate accident, not a property of the mechanism.)

Because the fake condition alters the label configuration as well as the reasoning, this reproduction fraction is not by itself a causal decomposition -- the matched controls below are.

\subsection{Matched Controls: Real Labels Only, and No Peers}

The labels-only condition shows exactly the label configuration the real round 1 showed -- same peers, same real verdicts -- with the reasoning withheld; the no-peers condition asks each judge to take a second careful look, bounding what no-peer reconsideration under the round-1 framing produces. On the common cohort of 280 candidates usable in all five conditions (Figure~\ref{fig:control}; full tables in Supplementary~B), the decomposition sharpens and delivers a surprise. Flip \emph{volume} is nearly identical everywhere (10.4--11.8\% of verdicts), and the no-peer second look raises unanimity from 43.2\% to 83.6\%: \textbf{79.6\% of the real jump, with no social signal whatsoever}. Random labels slow the endogenous convergence (78.2\%, below no peers) even as they steer flips toward whatever they display (89.5\%; Section 5.1); real labels complete 91.5\% of the jump. Adding the displayed reasoning context raises unanimity by a small but detectable 4.3pp: 18 candidates are unanimous only under full debate versus 6 only under labels-only (exact paired $p{=}0.023$; problem-clustered 95\% CI $[0.7,8.1]$pp). This is an interface-level increment, not a pure semantic-persuasion effect; most convergence is second-pass reconsideration aligned by displayed labels.

Seven contested-to-unanimously-wrong conversions occur in the paired sample; in the worked example of Supplementary~B, a dissenting judge that had correctly caught a bug capitulates with ``\emph{The majority of reviewers correctly identified this as a correct implementation}'' -- an appeal to the vote count, not to the code.

\section{Discussion}

\paragraph{Why post-debate intervention is a losing game.} Under \emph{substitution}, debate makes the tested execution-ballot intervention empirically inert after round 0 (Table~\ref{tab:rounds}); by round 2, with no pivotal cases, every single-ballot substitution is mechanically inert. Under \emph{override} -- the rule \citet{he2026minority} study -- inertness is replaced by risk: overturning must be aimed at decisions likely to be wrong, and one natural aiming signal, panel disagreement, is precisely what debate destroys (by round 3, every wrong decision in our data is unanimous, indistinguishable by vote pattern from the correct ones). An override policy facing a post-debate panel must find the errors with no help from the votes. Implementing simple policies on our data confirms this design concern (Supplementary~C): a margin-gated override is strictly corrective before debate ($+0.69$pp, every flip correct) but can never fire after round 0, while the ungated policy turns \emph{net negative} on post-debate votes (flip precision $1/3$, $-0.34$pp) -- the qualitative pattern behind their arbiter's 42.7\% flip precision and $-1.37$pp. Their setting differs in domain, judges, and rule, and we do not audit their runs; what our data establishes is that a useful disagreement-based targeting signal is gone before such a policy arrives, not that every learned arbiter must fail.

\paragraph{Design implications.} If one wants both debate and substitution-based verification, the measured opportunity lies \emph{before} any second-pass evaluation or peer exposure -- consult the verifier on round-0 margins; limiting the number of rounds is nearly useless (rounds 2--3 contribute only 7\% of the collapse), and even a peer-free second pass already destroys most disagreement. Second, post-debate unanimity should not be read as evidence of reliability: most of it would have formed with no peer input at all, substantially so under random labels, and debate destroys its own audit trail -- any disagreement-based uncertainty estimate, router, or abstention rule downstream of debate is reading a signal that debate has largely erased. Third, exploratorily, models show a stable descriptive ranking in susceptibility (order-of-magnitude spread, $\rho{=}0.99$; Table~\ref{tab:perjudge}), motivating broader, confound-controlled screening studies. Fourth, repeated parse failure is a cheap flag for contested candidates -- exactly the queries worth routing to a verifier or a human.

\paragraph{Limitations.} First, attrition: we verified it is non-random, controlled for it with paired samples, and showed the collapse survives excluding the dominant failure source -- but the fixed cohort still over-represents easy candidates, and heavily contested candidates remain unobservable \emph{inside} debate while judges' outputs on them cannot be parsed. The failures are concentrated by source (Doubao-Seed-1.6 is involved in 256 of the 277 round-1 drops; full attribution in Supplementary~A), so the ``formatting degrades under contention'' observation is largely driven by the most failure-prone members. Second, the controls identify interface-level increments rather than semantic persuasion in isolation: full debate versus labels only also changes context length, formatting, and attention allocation; only the first 400 characters of each peer rationale are displayed, and the 400-token generation cap may suppress extended argument. Length-matched and shuffled-rationale controls remain future work; the fake condition additionally uses two fixed templates and unmatched label configurations. Third, our primary results are from code correctness under one debate protocol at temperature 0. Two additional real round-1 runs and two additional fake-peer seeds reproduce the one-round effects on paired cohorts (real unanimity $94.0$--$95.3\%$; fake $73.9$--$77.3\%$; Supplementary~F), and a smaller MATH-500 study provides cross-domain corroboration -- one round takes unanimity from 73.8\% to 97.2\%, eliminates all ten pivotal candidates, and grows unanimously wrong verdicts from 1 to 7 ($n{=}141$; Supplementary~G) -- but it does not repeat every control and domain breadth remains limited (panel sizes 3--7 and leave-one-judge-out compositions: Supplementary~E). Fourth, our direct round-by-round measurements concern substitution; Supplementary~C implements simple margin-gated and ungated execution-based overrides, but we do not reproduce a learned LLM arbiter. Fifth, the round-0 pivotal gain is directionally positive but not significant at our scale, and the mild accuracy gain under fake verdicts is a base-rate accident, not evidence that conformity is benign.

\paragraph{Reproducibility.} Primary code-domain quantities derive from stored model-output artifacts (blind verdicts and execution ground truth; the four-round log; additional real and fake runs; labels-only; and no-peer), while the MATH corroboration adds its own logs. Stored debate records contain verdicts and the displayed response excerpts rather than unrestricted full API responses. The 600-candidate batch is a seeded shuffle (seed 42); fake-verdict draws use recorded random streams. Judges run at temperature 0 with a 400-token cap (more than $10^4$ calls, retries included); analyses run in Python (NumPy/SciPy) on a commodity CPU. We release the stored records, prompts, and deterministic analysis scripts. All reported statistics are recomputed from those records; Figure~\ref{fig:teaser} is a schematic using rounded, cross-checked values.

\section{Conclusion}

Interventions on a debate panel often rely on disagreement, and debate destroys it almost entirely within one round: by round 3 every wrong decision is unanimous, and a substituted verification ballot with real corrective effect before debate has none after it. Controls show that second-pass reconsideration and displayed-label alignment explain most of the collapse; the full-debate condition exceeds labels only by a small residual under this interface. The verifier was never useless -- the debate got to the disagreement first.

\footnotesize
\setlength{\bibsep}{0pt}
\bibliographystyle{plainnat}
\bibliography{references}

\end{document}